\documentclass[conference]{worldcomp}

\usepackage[hmargin=.75in,vmargin=1in]{geometry}
\usepackage[american]{babel}
\usepackage[T1]{fontenc}
\usepackage{times}
\usepackage{caption,url}

\usepackage{textcomp}
\usepackage{epsfig,graphicx}
\usepackage{xcolor}
\usepackage{amsfonts,amsmath,amssymb,tabularx}
\usepackage{fixltx2e} % Fixing numbering problem when using figure/table* 
\usepackage{booktabs}

\title{\bf Information-theoretic Interestingness Measures for Cross-Ontology Data Mining}          

\author{
{\bfseries Prashanti Manda$^{1*}$, and Fiona McCarthy$^2$, and Bindu Nanduri$^3$, and Hui Wang$^3$, and Susan M. Bridges$^4$}\\
$^1$Department of Biology, University of North Carolina at Chapel Hill, Chapel Hill, NC, USA\\
$^2$Department of Veterinary Science and Microbiology, University of Arizona, Tucson, AZ, USA\\
$^3$Department of Basic Sciences, College of Veterinary Medicine, Mississippi State University, MS, USA\\
$^4$Information Technology and Systems Center, University of Alabama, Huntsville, AL, USA\\
$^*$Corresponding author
}

\begin{document}

\maketitle                        %%%% To set Title and Author names.

\begin{abstract}
Community annotation of biological entities with concepts from multiple bio-ontologies has created large and growing repositories of ontology-based annotation data with embedded implicit relationships among orthogonal ontologies.  Development of efficient data mining methods and metrics to mine and assess the quality of the mined relationships has not kept pace with the growth of annotation data. In this study, we present a data mining method that uses ontology-guided generalization to discover relationships across ontologies along with a new interestingness metric based on information theory. We apply our data mining algorithm and interestingness measures to datasets from the Gene Expression Database at the Mouse Genome Informatics as a preliminary proof of concept to mine relationships between developmental stages in the mouse anatomy ontology and Gene Ontology concepts (biological process, molecular function and cellular component). In addition, we present a comparison of our interestingness metric to four existing metrics.
Ontology-based annotation datasets provide a valuable resource for discovery of relationships across ontologies. The use of efficient data mining methods and appropriate interestingness metrics enables the identification of high quality relationships.
\end{abstract}

\vspace{1em}
\section{Introduction}
% no \PARstart

The wide spread use of ontologies to describe data has led to the availability of large ontology-based datasets where different ontologies are often used to describe distinct characteristics of entities. For example, in the biological and bio-medical domain, the Gene Ontology might be used to describe the biological processes of a gene product while an anatomy ontology is used to specify the location of expression. The integration of these distinct ontology-based datasets lends itself to the discovery of interesting relationships between the ontologies (cross-ontology relationships). These relationships enable data and information integration and lead to the discovery of patterns not evident from individual datasets. For example, cross-ontology relationships mined from gene expression and annotation data can be used to answer ``big picture'' questions such as ``What biological processes are typically expressed in the mouse brain?''

The abundance of  ontology-based annotations is accompanied by a dearth of efficient data mining techniques that can discover biologically relevant relationships from the data. One of the drawbacks of data mining techniques such as association rule mining is the retrieval of large number of relationships that need to  be prioritized or ranked based on domain knowledge. Existing ranking metrics are either unsuitable for ontology based relationships \cite{prashanti2} or do not accommodate domain knowledge. The gap in techniques to mine and rank biological ontology-based relationships forms the motivation for this work.  

This paper focuses on data mining methods for the integration and mining of ontology-based annotation datasets and describes a new information theoretic metric to rank the mined cross-ontology relationships (relationships between concepts from different ontologies). Our data mining algorithm integrates ontological datasets and mines cross-ontology association rules to indicate the relationships between two ontologies describing different aspects of biological entities.  Note that this form of discovery is distinct from efforts to map concepts across different ontologies describing the same aspects of entities. 

An association rule is defined as an implication of the form $x \rightarrow y$ where $x$ (antecedent) and $y$ (consequent) are co-occurring items derived from a transaction set $T$. In association rules that describe market sales, transactions are sets of items purchased together. In cross-ontology association rules derived from annotation data, each transaction contains a gene name and one or more annotations from each ontology (in this study, GO and anatomy ontologies) and $x$ and $y$ are co-annotated concepts from different ontologies. In the data used for this study, transactions express the involvement of gene products in specific processes/functions/components and expression of the gene product in specific tissues. Additionally, $x$ and $y$ are restricted to single concepts instead of itemsets. 

Our data mining algorithm employs subsumption reasoning to mine relationships at multiple levels across the input ontologies. Our previous work on generalization algorithms explored two methods of generalization: 
\begin {enumerate}
\item Level-by-level generalization \cite{prashanti}.
Depth of annotations is used to conduct incremental generalization and mining one level at a time. 
 
\item Generalization to all ancestors via transitive relationships \cite{prashanti2}.
This generalization method is an improvement over the level-by-level generalization since it does not rely on the depth of annotations as a guide for generalization. Instead, generalization is conducted in a single step where all annotations are supplemented with all their ancestors in the ontology. The mining step is conducted only once after the generalization process to improve efficiency.
\end{enumerate}

These algorithms have been applied to GO annotation data and were used to discover relationships across the ontologies of the GO. While our previous methods use the depth of ontology terms to guide generalization, Information content has been shown to be a more accurate indicator of ontology term specificity as compared to depth in the ontology \cite{gopad,alterovitz}. This is because ontologies evolve over time and different sections of an ontology are developed to different extents depending on the level of available scientific knowledge and the involvement of the specific research community. In the research reported in this paper, we propose a new information theoretic interestingness measure called Integrated Rule Information Content ($IRIC$) to inform ontology-enabled association rule mining from multiple ontologies. $IRIC$ combines the information content of the terms in a rule with the shared information among the terms to accurately assess the interestingness of the rule. $IRIC$ is calculated from the following two components:
\begin{enumerate}
\item	Normalized Information Content ($N\_IC$): $N\_IC$ indicates the information content of ontology terms in a cross-ontology association rule.
\item	Normalized Cross-ontology Mutual Information ($N\_COMI$): $N\_COMI$ quantifies the information shared by the terms in a cross-ontology association rule.
\end{enumerate}

We apply our data mining algorithm to GO annotation and tissue expression data from the Gene Expression Database at MGI \cite{mgi} to discover relationships between the GO ontologies and the Mouse Anatomy ontology. $N\_IC$ and $N\_COMI$ thresholds are used to filter uninformative terms and relationships while $IRIC$ scores are used to rank the remaining relationships.

\section{Related Work}
Association rule mining has been applied to ontology based mining by several previous studies to discover relationships between one or multiple ontologies \cite{faria2012mining, integratedanalysis, secondgo, nonlexical, benites2014mining, ferraz2013ontology, paul2014semantic} . In the majority of these studies, relationships are discovered from a single ontology \cite{faria2012mining, integratedanalysis, secondgo, nonlexical, ferraz2013ontology, paul2014semantic} while some methods can be used for cross-ontology mining as well \cite{benites2014mining}. 

While association rule mining in annotation datasets has been explored widely, few studies have focused on developing alternative and appropriate interestingness metrics for ontology based association rules \cite{faria2012mining, benites2014mining,paul2014semantic}. Faria et al. use association rules to quantify annotation inconsistency by exploring erroneous, incomplete, and inconsistent annotations. Although, Support and Confidence are used as initial interestingness metrics during mining, Faria et al. employ strategies post mining to filter the discovered rules. The metrics (Generic rules, Agreement, Ancestral and descendant distance) use ontology semantics to weed out uninteresting rules. These metrics are similar to the post filtering techniques we have used in our previous work \cite{prashanti2}. 

Another notable work in this area, Benites et al., proposes the idea of comparing the real value of a rule's interestingness with the expected value \cite{benites2014mining}. Rules with more significant differences in these values are considered more rare and interesting. Paul et al. introduce a suite of metrics based on semantic similarity and ontological distance adapted from existing metrics. These metrics are applied to discover relationships between human phenotypes (HPO terms) and bone dysplasias. While relationships with semantically similar terms are more valuable in Paul et al.'s application, the ontologies we are using for cross-ontology relationships capture different aspects of the objects and need not be semantically similar to be interesting \cite{paul2014semantic}.

\section{Materials and Methods}
This section describes the generalization method and information theoretic interestingness metric we developed for cross-ontology data mining. 

\subsection{Generalization and mining}
As a preprocessing step for the mining algorithm, gene annotations from different ontologies (in this case, anatomy and GO) are combined to build a transaction set. Each transaction in this set contains a gene along with GO and anatomy annotations for the gene.  We apply our generalization algorithm to simultaneously generalize terms from all of the ontologies represented in the transaction set \cite{prashanti2}. Generalization supplements the annotations in a transaction with all of their ancestors related via transitive relations. The generalized transactions are then processed to remove uninformative terms using an $N\_IC$ threshold as described in Section \ref{icthresholds}. The resulting generalized transactions are mined using Christian Borgelt's implementation of the Apriori algorithm \cite{apriori}. The mined relationships are further filtered using a $N\_COMI$ threshold to remove relationships with insufficient shared information. $IRIC$ is then used to rank the remaining relationships.  

\subsection{Integrated Rule Information Content}
Integrated Rule Information Content ($IRIC$) is a novel interestingness measure that combines information content ($N\_IC$) of concepts in a rule with the shared information in the rule ($N\_COMI$). $IRIC$ of a rule $x \rightarrow y$ is defined as 

\begin{flalign*}
& IRIC_{x \rightarrow y} = ((\alpha * N\_IC_x) + (\beta * N\_IC_y)) * N\_COMI_{x \rightarrow y} & \\
& \textrm{where } \alpha \textrm{ , } \beta \textrm{ are weighted coefficients of concepts} & \\
& \textrm{from the ontologies of $x$ and $y$, and } & \\
& \alpha + \beta = 1 &
\end{flalign*}

Concepts from both the ontologies can be weighted equally by setting $\alpha$ and $\beta$ to 0.5. Alternatively, greater or lower weights can be attributed to concepts from one ontology by modifying $\alpha$ or $\beta$. The range of the $IRIC$ measure is [0, 1].
The components used to calculate $IRIC$ are defined below. 

\subsubsection{Information content of concepts ($N\_IC$)} 
 
We define Normalized Information Content ($N\_IC$) of a term $t$ as

\begin{flalign*}
& {N\_IC}_t =\frac{-log p(t)}{UB(IC)} \textrm{ where ;} & \\
& p(t)=\frac{|G_t| +\displaystyle\sum\limits_{i=1}^j |G_{C_i}|}{|G|} \textrm{ and ;} & \\ 
& G = \textrm{set of all genes in the transaction set,} & \\
& G_t = \textrm{set of genes annotated to t,} & \\
& C_i =\{1,2, \cdots, j\} \textrm{ are the descendants of t in the ontology,}. & \\ 
& G_{C_i} =\textrm{set of genes annotated to descendant $C_i$} \\
& \textrm{Upper bound for IC, } UB(IC) = -log(\frac{1}{|G|}) & \\
\end{flalign*}

Our definition of $N\_IC$ is adapted from Shannon's information content \cite{shannon} to take into account the implicit annotations indicated by subsumption reasoning over the ontology and to make $N\_IC$ comparable to other metrics by restricting the range of the metric to [0,1].

\paragraph{Cross-ontology Mutual Information}
Mutual information of an association rule captures the shared information content and inter-dependence of the antecedent and the consequent in the rule. The mutual information (MI) \cite{mutual} of an association rule  $x \rightarrow y$ , where $x$ and $y$ are items from the transaction set, is defined as 
\begin{equation*}
MI = p(xy) * log_2(\frac{p(xy)}{p(x)p(y)})
\end{equation*}
This definition of MI uses the entire set of transactions as the background to compute the probabilities thus assuming that all transactions contain annotations from all ontologies in the analysis. However many biological datasets incur the problem of missing data where entities are not annotated to all ontologies in the analysis \cite{prashanti2}. To address this issue of missing data, we adapted the standard definition of MI to define Normalized Cross-ontology Mutual Information ($N\_COMI$) for assessing the interestingness of cross-ontology multi-level association rules.

We use the following sets in the definition of Normalized Cross-ontology Mutual Information where $x \rightarrow y$
represents a cross-ontology rule with $x$ and $y$ belonging to different ontologies. Note that these sets are subsets of the input transaction set. 
\begin{enumerate}

\item	$X_{(x \rightarrow y)}$ is the set of transactions containing $x$ and at least one term from the ontology of $y$. 
\item	$Y_{(x \rightarrow y)}$ is the set of transactions containing $y$ and at least one term from the ontology of $x$. 
\item	$COCategory_{x \rightarrow y}$ is the set of transactions containing at least one term from $x$'s ontology and and one from 
$y$'s ontology. 
\item $XY_{x \rightarrow y}$ is the set of transactions which contains both $x$ and $y$.
  \end{enumerate}
The normalized cross-ontology mutual information ($N\_COMI$) of a rule, $x \rightarrow y$ is defined as

\begin{flalign*}
& N\_COMI_{x \rightarrow y} = \frac{p(xy) * log_2 \frac{p(xy)}{p(x)p(y)}}{min((-log_2p(x)p(x)),-log_2p(y)p(y)))} \textrm{ with;} & \\
& p_x = \frac{|X_{x \rightarrow y}|}{|COCategory_{x \rightarrow y}|} \textrm{ ,} & \\
& p_y = \frac{|Y_{x \rightarrow y}|}{|COCategory_{x \rightarrow y}|} \textrm{ and,} & \\
& p_{xy} = \frac{|XY_{x \rightarrow y}|}{|COCategory_{x \rightarrow y}|} & 
\end{flalign*}

\subsection{$N\_IC$ and $N\_COMI$ thresholds}
\label{icthresholds}
First, uninformative ontology terms are removed from the transaction set after generalization and prior to mining using an $N\_IC$ threshold.  This step helps avoid mining rules with uninformative terms that occur frequently in the transaction set. These terms are typically closer to the root of the ontology. In our analysis, uninformative terms are terms that are annotated to many genes in the dataset. Examples of uninformative terms in our dataset include  organ system and nervous system in the anatomy ontology,  \textit{GO:0005623} (cell), \textit{GO:0065007}	(biological regulation), and \textit{GO:0005488} (binding) from the Gene  Ontology. Selecting an $N\_IC$ threshold is a subjective choice and depends on the application of the discovered rules, the ontologies in question, and the annotation dataset.

Second, an $N\_COMI$ threshold is selected using Monte Carlo methods. A synthetic dataset containing the same number of transactions as the transaction set is generated using sampling with replacement from the set of all terms in the transaction set. Cross-ontology multi-level rules are mined from the synthetic data and the $N\_COMI$ of the rules is calculated. The rules mined from the synthetic data are considered to be False Positives while rules mined from the actual transaction set are `True Positives'. The False Positives and True Positives are combined and rules are ranked by $N\_COMI$.  A $N\_COMI$ threshold is selected to yield a desired false positive rate. This $N\_COMI$ threshold is used to eliminate uninteresting rules mined from the actual transaction set.

$N\_IC$ and $N\_COMI$ are both necessary because they capture different properties of the rules. $N\_IC$ represents the specificity of terms in the rules while $N\_COMI$ captures the information shared by the antecedent and consequent.  Our goal is to mine rules with highly informative terms where the rule mutual information is also high.  The dual application of $N\_IC$ and $N\_COMI$ thresholds removes terms with little information and leads to the discovery of rules with high mutual information content. 

\subsection{Properties of Cross-ontology Mutual Information ($N\_COMI$) and Integrated Rule Information Content ($IRIC$)}
The $N\_IC$ of a concept $t$ is 0 (lowest) when $p(t) =1$ and is 1 (highest) when $t$ occurs only once in the transaction dataset. The $N\_COMI$ of a rule is 0 when the concepts in the rule are statistically independent. The $IRIC$ of a rule is 0 (lowest) when the $IC$ of both the concepts in the rule and the $N\_COMI$ of the rule are 0. The $IRIC$ is 1 (highest) when the $IC$ of both the concepts in the rule and the $N\_COMI$ of the rule are 1.
 Tan et al. identify three key properties of a desirable metric  (\cite{rightinterestingness} ): An interestingness metric, M is considered desirable if it satisfies the following three properties for a rule of the form $x \rightarrow y$. 
 \begin{enumerate}
 
\item M is 0 when $x$ and $y$ are statistically independent.
\item M monotonically increases with $p(xy)$ when $p(x)$ and $p(y)$ remain the same. $p(x)$, $p(y)$, and $p(xy)$ are the probabilities of observing x, y, or both in a transaction respectively.
\item M monotonically increases with $p(x)$ or $p(y)$ when the rest of the parameters remain the same. 
\end{enumerate}

These properties are meant to be applicable for metrics that quantify association rules and not individual terms. In our analysis, the metrics we use to quantify association rules are $N\_COMI$ and $IRIC$ while $N\_IC$ is used to quantify informativeness of terms. We list the behavior of $N\_COMI$ and $IRIC$ with respect to these properties in Table~\ref{tab:properties}.

\section{Main Results}
We designed an experiment as a preliminary proof of concept to demonstrate the mining and ranking of cross-ontology relationships using the $IRIC$ metric. The data used for this experiment was gene expression data in post-natal mouse from the Gene Expression Database (GXD) \cite{gxd} at the Mouse Genome Informatics (MGI). The transaction set built from this data contains 8,176 transactions and 123,069 GO terms and 124,920 anatomy terms. Each transaction contains a gene product name accompanied by one or more annotations to the anatomy and gene ontologies.

Cross-ontology rules were mined after generalization and the $N\_IC$ and $N\_COMI$ information theoretic metric thresholds were applied incrementally. The $IRIC$ metric, a combination of  $N\_COMI$ and  $N\_IC$ was used to rank the remaining mined rules after the thresholds were applied. In this experiment, we weighted GO and Anatomy concepts equally by setting $\alpha$ and $\beta$ to 0.5.

\subsection{Effect of $N\_IC$ and $N\_COMI$ thresholds}
For this experiment, we chose to only include GO and Anatomy terms that were annotated to no more than 5\% of the total genes in the dataset. This threshold was selected empirically. This translates into an $IC$ of 4.32 (same for any dataset) and $N\_IC$ of 0.33 (specific to our dataset). The percentage of annotated genes in computing the $N\_IC$ threshold can be varied depending on the level of informativeness desired in the relationships. The greater the percentage of genes annotated to a term, the lower the $N\_IC$ of the term. A practical consideration in choosing this threshold is to explore the distribution of $N\_IC$ scores in the data and select a threshold that balances the number of terms for analysis and the information content of the terms. Different choices of percent genes annotated to a term and how this translates to $IC$ (dataset independent) and $N\_IC$ (specific to our study) are shown in Table \ref{tab:icthreshold}. The Monte Carlo method described in Section \ref{icthresholds} was used to select a threshold for $N\_COMI$.  The selected $N\_COMI$ threshold was used to remove uninformative rules.

Table~\ref{tab:table2} provides a summary of the experimental results and shows the effect of  $N\_IC$ and $N\_COMI$ in the mining process.  We measure the effect of each of these components with respect to the number of rules mined, the average $N\_IC$, and the average $N\_COMI$.

When an $N\_IC$ threshold was applied alone, (Table~\ref{tab:table2}, column 3), 91.16\% of the mined rules are removed as uninteresting,  the average $N\_IC$ increases by approximately 96\%  and the average $N\_COMI$  increases by approximately 55\%. 

When the $N\_COMI$ threshold is applied alone (without the $N\_IC$ threshold, Table~\ref{tab:table2} column 4) there is a much smaller (2\%) reduction in the number of rules than seen with the $N\_IC$ cutoff. The average $N\_COMI$ of rules increases by 3.44\%. 

The last column in Table~\ref{tab:table2} demonstrates the synergistic effects of using both $N\_IC$ and $N\_COMI$ thresholds. Both the average $N\_IC$ and $N\_COMI$ scores are at the highest when both thresholds are applied together as compared to singular application of either one of the thresholds. These results demonstrate that the combined application of $N\_IC$ and $N\_COMI$ thresholds removes uninteresting rules effectively resulting in rules with high mutual information and containing informative terms.

\subsection{Evaluation of the $IRIC$ metric}
The $IRIC$ metric was evaluated by comparing it to a commonly used information theoretic measure, Information Gain \cite{lenca2008selecting}. The top 100 rules ranked by $IRIC$ were manually compared by a biologist to those ranked by Information Gain for evidence in published literature. The biologist attempted to validate each rule by conducting literature searches for evidence of a relationship between the antecedent and the consequent of the rule. If evidence of such a relationship was found in literature, the corresponding rule was categorized as "Validated" and the provenance information of the literature was recorded. If no evidence was found, the rule was marked as "Not Validated". Additionally, each rule was evaluated for its meaningfulness and categorized as either "Meaningful" or "Not meaningful". The meaningfulness of a rule indicates whether or not it makes sense for the items in the rule to be co-annotated \cite{prashanti}.  The definitions of these categorizations are as per our previous work on cross-ontology association rule mining \cite{prashanti}. This evaluation was conducted based upon the biologist's personal, biological knowledge, and literature searches.

Of the top 100 $IRIC$ rules, literature-based evidence was found for 92 rules (S1 File). In contrast, evidence was found for only 78 of the top Information Gain rules. One $IRIC$ rule was categorized as "Not meaningful" while six Information Gain rules were categorized as "Not meaningful". These preliminary results show that $IRIC$'s top ranked rules have a high accuracy rate and that $IRIC$ outperforms Information Gain in the percentage of manually validated and meaningful rules. These evaluated rules and the results of manual validation are available publicly at  (\url{https://github.com/prashanti/Supplementary_Files}).

\section{Conclusions}
The widespread use of ontologies to represent data and knowledge has led to the availability of vast amounts of ontology-annotated data. However, there is a dearth of efficient algorithms and ontology-aware metrics to mine multi-ontology data and rank the mined relationships.  In this study, we developed information theoretic metrics to rank cross-ontology rules. We presented the use of our mining algorithm along with the metrics to mine relationships across the Gene Ontology and Anatomy Ontology. Our results demonstrate that our proposed metric, $IRIC$ is effective at ranking accurate relationships mined from annotation data.
\subsection{Tables}

\begin{table}[h]
\caption{Properties satisfied by the information theoretic measures Cross-ontology Mutual Information ($N\_COMI$) and Integrated Rule Information Content ($IRIC$).}
\begin{center}
\begin{tabularx}{\linewidth}{XXX}
Property	& $N\_COMI$	& $IRIC$ \\\hline
1	& Satisfies & Satisfies \\
2	& Satisfies	& Satisfies \\
3	& Does not satisfy & Does not satisfy
\end{tabularx}
\label{tab:properties}
\end{center}
\end{table}

\begin{table}[h]
\caption{$IC$ and $N\_IC$ thresholds corresponding to percentage of gene annotations to terms.}
\begin{center}
\begin{tabularx}{\linewidth}{XXX}
 \hline
 Threshold of \% genes annotated to a term. & $IC$ & $N\_IC$ (Number of genes = 8176) \\ \hline

25\% & 2.25 & 0.17 \\
20\% & 2.32 & 0.18  \\
15\% & 2.73 & 0.21  \\
10\% & 3.32 & 0.26 \\
5\% & 4.32 & 0.33  \\
4\% & 4.64 & 0.36 \\
3\% & 5.05 & 0.39  \\
2\% & 5.64 & 0.43 \\
1\% & 6.64 & 0.51 \\

\end{tabularx}
\label{tab:icthreshold}
\end{center}
\end{table}

\begin{table}[h]
\caption{Comparison of the number of rules mined, average $N\_IC$, and average $N\_COMI$ when $N\_IC$ and $N\_COMI$ thresholds are applied individually and together.}

\begin{center}
\begin{tabularx}{\linewidth}{XXXXX}
    \hline
    	 &	Before pruning & 	Only $ N\_IC$  threshold applied & Only	$N\_COMI$ threshold applied &	Both $N\_IC$ and $N\_COMI$ thresholds applied \\
        \hline
    Number of rules mined &	66437	& 5873 &	64908 &	4925\\
   Average $N\_IC$ &	0.28&	0.55&	0.33&	0.55\\
   Average $N\_COMI$ &	0.058&	0.13	&0.06&	0.148\\
   
\end{tabularx}
\label{tab:table2}
\end{center}
\end{table}

\clearpage

\bibliographystyle{IEEEtran.bst}
\bibliography{References}

\end{document}